\documentclass[11pt]{article}
\usepackage{coling2018}
\usepackage{times}
\usepackage{url}
\usepackage{latexsym}

\usepackage{graphicx}
\usepackage{amsmath}
\usepackage{multirow}
\usepackage{diagbox}
\usepackage{booktabs}
\usepackage[normalem]{ulem}
\usepackage{subcaption}
\usepackage{wrapfig}
\usepackage{lipsum}
\date{}
\title{Double Path Networks for Sequence to Sequence Learning}

\author{
	$^1$Kaitao Song, $^2$Xu Tan, $^3$Di He, $^1$Jianfeng Lu, $^2$Tao Qin and $^2$Tie-Yan Liu \\
	$^1$Nanjing University of Science and Technology \\
	$^2$Microsoft Research \\
	$^3$Key Laboratory of Machine Perception, MOE, School of EECS, Peking University \\
	{\tt \{kt.song, lujf\}@njust.edu.cn} \\
	{\tt \{xuta, taoqin, tyliu\}@microsoft.com} \\
	{\tt \{di\_he\}@pku.edu.cn}
}

\date{}

\begin{document}
\maketitle

\begin{abstract}
	Encoder-decoder based Sequence to Sequence learning (S2S) has made remarkable progress in recent years. Different network architectures have been used in the encoder/decoder. Among them, Convolutional Neural Networks (CNN) and Self Attention Networks (SAN) are the prominent ones. The two architectures achieve similar performances but use very different ways to encode and decode context: CNN use convolutional layers to focus on the local connectivity of the sequence, while SAN uses self-attention layers to focus on global semantics. In this work we propose Double Path Networks for Sequence to Sequence learning (DPN-S2S), which leverage the advantages of both models by using double path information fusion. During the encoding step, we develop a double path architecture to maintain the information coming from different paths with convolutional layers and self-attention layers separately. To effectively use the encoded context, we develop a cross attention module with gating and use it to automatically pick up the information needed during the decoding step. By deeply integrating the two paths with cross attention, both types of information are combined and well exploited. Experiments show that our proposed method can significantly improve the performance of sequence to sequence learning over state-of-the-art systems.
\end{abstract}

\section{Introduction}
	\blfootnote{
		%
		%
		%
		\hspace{-0.65cm}
		This work was done when K. Song was an Intern at Microsoft Research \\
		Our work is built upon \url{https://github.com/StillKeepTry/Transformer-PyTorch} based on fairseq. \\
		\hspace{-0.65cm}  
		This work is licensed under a Creative Commons 
		Attribution 4.0 International License.
		License details:
		\url{http://creativecommons.org/licenses/by/4.0/}.
	}

	Sequence to Sequence learning (S2S) \cite{DBLP:conf/emnlp/ChoMGBBSB14,DBLP:conf/nips/SutskeverVL14} is one of the challenging tasks in artificial intelligence, which covers machine translation, document summarization and question answering, etc. It has attracted more and more attention in recent years due to the success of deep learning. Sequence to sequence learning is usually developed based on a general encoder-decoder framework: The encoder reads the source sequence and generates a set of representations. After that, the decoder estimates the conditional probability of each target token given the source representations and its preceding tokens. 
	
	Different network architectures for sequence to sequence modelling have been designed based on this framework. Long Short-Term Memory (LSTM) based model \cite{DBLP:conf/emnlp/ChoMGBBSB14} is the most popular. In the LSTM based model, the tokens are encoded/decoded using LSTM units, which can effectively summarize the temporal information of the sentences in source/target domains. However, because of its recurrent nature, the LSTM based model is difficult to parallel, so the training efficiency becomes the major challenge. Recently, Convolutional Neural Network (CNN) based \cite{DBLP:conf/icml/GehringAGYD17} and Self Attention Network (SAN) based \cite{NIPS2017_7181} models have been proposed, which achieved significant improvements in accuracy. Both models replace the recurrent structure with networks that are easy to parallelize, leading to training time acceleration.
	
	CNN based and SAN based sequence to sequence models encode and decode information in very different ways. CNN based model treats the sequences similar with images: The convolutional layer is used to summarize information for a local region of the sequence, which corresponds to a subsequence of tokens. Different from CNN based model, SAN based model uses self-attention layer to abstract the context based on semantic meanings: In each layer, for each token, the similarity between the token and all other tokens in the sequence are computed. Then the contexts are weighted combined based on the global semantic similarity. By stacking multiple convolutional layers or self-attention layers, the two models obtain the whole contextual information during the encoding and decoding step.
	
	As we can see, the convolutional layer focuses more on abstracting information according to local connectivity, while self-attention layer focuses more on using global semantics. Both models achieve similar accuracy on several sequence to sequence learning tasks. Then a natural question raises: is there any way to integrate the two methods together to achieve a higher accuracy? This is exactly the purpose of the paper.
	
	In this work, we investigate the possibility to combine the two models into a unique framework and develop efficient architecture to leverage both advantages. In particular, we propose Double Path Networks for Sequence to Sequence learning (DPN-S2S), which contain a convolutional path and a self-attention path with attention information fusion between the encoder and the decoder. In the encoding step, we use the convolutional path and self-attention path independently. Thus we can summarize the context of the source sequences from different aspects. In the decoding step, we develop a cross attention module with gating to automatically select appropriate context, and use the context in a similar double path structure. As there is no recurrent structure in convolutional layers and self-attention layers, the model is still easy to parallelize. Since we abstract the hidden representation from different ways, the contexts we obtain are more diverse, making it possible to achieve higher accuracy under the same parameter number. 
	
	Experimental results show that our proposed approach can achieve significant improvements than single path based S2S framework. The results are summarized as follows: 
	
	\begin{itemize}
		\item With a similar number of parameters, our model outperforms CNN based, SAN based model. More specifically, on Nist Chinese-English translation task, we outperform the strong CNN based baseline by 0.46-2.96 BLEU, SAN based baseline by 0.74-0.99 BLEU respectively in different Nist test sets. Furthermore, our model also achieves state-of-art BLEU/ROUGE score on IWSLT14 German-English translation task and English Gigaword text summarization task. 
		\item Through ablation studies and analyses we show that both the CNN and SAN path are important in the encoder and the decoder. Removing any part would cause performance drop, which demonstrates the effectiveness of our model.  
	\end{itemize}
	
\section{Background}
	\subsection{Encoder-Decoder Framework}
		Denote $x=(x_1,x_2,...,x_{m})$ and $y=(y_1,y_2,...,y_{n})$ as the sentence pair, in which $x_i$ and $y_t$ are the $i$-th and $t$-th tokens for sequences $x$ and $y$, $m$ and $n$ are the lengths of the sequences. The sequence to sequence model learns to estimate the conditional probability $ P(y|x)$. Denote the model parameter as $\theta$ and denote the training corpus as $S$. One of the most popularly used objective functions is the log likelihood function:

		\begin{equation}
			L(\theta;S)=\frac{1}{|S|}\sum_{(x,y)\in S}\log P(y|x; \theta). 
		\end{equation}
		According to the probability chain rule, we have:
		\begin{equation}
			P(y|x; \theta) = \prod_{t=1}^{n} P(y_{t}|y_{<t},x;\theta),
		\end{equation} 
		where $y_{<t}$ is the sequence of proceeding tokens before position $t$. To model the conditional probability, an encoder-decoder framework is usually employed \cite{DBLP:conf/emnlp/ChoMGBBSB14,DBLP:journals/corr/BahdanauCB14}. In the encoding phase, $(h_1,h_2,\cdots,h_{m})$ is learnt as the representations of the source sequence:
		\begin{eqnarray}
			(h_1,h_2,\cdots,h_{m})=f(x_1,x_2,...,x_{m}).
		\end{eqnarray}
		where $f(\cdot)$ can be implemented as RNN, CNN or SAN. During the decoding phase, the decoder generates the target token $y_t$ at position $t$ using the previous generated tokens $y_{<t}$ and the source representations $(h_1,h_2,\cdots,h_{m})$.

		An attention mechanism is usually adopted to choose the places in the source representations to focus on. There are several types of attention used in the literature, such as $dot$, $concat$ and $general$ \cite{DBLP:conf/emnlp/LuongPM15,DBLP:journals/corr/BahdanauCB14}. In this paper, we mainly adopt the $dot$ type which is formulated as the q-k-v form:

		\begin{equation}
			\label{eq:eq3}
			Attention(q,k,v) = softmax(qk^T) v,
		\end{equation}
		where $softmax$ function computes the attention coefficient which evaluates how well the source representations $k$ matches the target query $q$. $q$ is usually the hidden state in the current decoder layer and $k$, $v$ are usually the same as source representation $h$.

		\subsection{CNN based Encoder-Decoder}
		CNN based Encoder-Decoder framework \cite{DBLP:journals/corr/GehringAGYD17} typically uses one-dimensional convolution in each layer to perform nonlinear transformation. 
		Denote $h_{i}^l$ with dimension $d$ as the $i$-th hidden state in the $l$-th CNN layer and set $h_{i}^0=e_i$, where $e_i$ is the input embedding of the sequence. The hidden state $h_{i}^l$ is computed by the convolution operation:
		\begin{equation}
			\label{eq:convolution}
			h_i^{l} =v([h_{i-r/2}^{l-1}, ... ,h_{i+r/2}^{l-1}] W^l   + b_w^l) + h_i^{l-1},
		\end{equation}
		where $W^l \in \mathcal{R}^{rd \times 2d}$ is the convolution filter with kernel size $r$, $b_w^l \in \mathcal{R}^{2d}$ is the bias and $d$ is the hidden dimension. The convolution input is the concatenation of $r$ elements in the lower layer with hidden dimension $d$. We use Gated Linear Units (GLU) \cite{DBLP:conf/icml/DauphinFAG17} as the activation function $v$, with $2d$ input vector and $d$ output vector.

	\subsection{SAN based Encoder-Decoder}
		SAN system is first proposed by \newcite{NIPS2017_7181}. Each single self-attention layer has two sublayers: a multi-head self-attention layer and a feed forward network. Both sublayers are stacked using residual connection \cite{DBLP:conf/cvpr/HeZRS16} and layer normalization \cite{DBLP:journals/corr/BaKH16}. The multi-head self-attention layer $MH(\cdot)$ is formulated as follows:
		\begin{equation}
			\begin{aligned}
			MH(q,k,v) &= {Concat}_{i=1}^s h_i(q,k,v), \\
			h_i(q,k,v) &= Att(\frac{qW_i^q}{\sqrt{d_s}}, kW_i^k, vW_i^v). \\
			\end{aligned}
		\end{equation}
		Each head uses weights $W_i^{q}$, $W_i^{k}$ and $W_i^{v}$ to linearly transform the input $q$, $k$, $v$ and then performs $Att(\cdot)$ which is equal to Equation \ref{eq:eq3}. $W_i^{q}$, $W_i^{k}$ and $W_i^{v} \in \mathcal{R}^{d \times d_{s}}$, where $d_{s}$ is a scale factor, which equal to $d/s$, $d$ is the hidden size of q, and $s$ is the number of heads.   	

		For the feed forward network, we use a simple two-layer, fully connected network with ReLU activation function in the middle layer: 
		\begin{equation}
			\begin{aligned}
				FeedForward(x) = f_2(Max(0, f_1(x))), 
			\end{aligned}
		\end{equation}
		where $f_1(\cdot)$ and $f_2(\cdot)$ are both feed forward network. The feed forward network applies the non-linear transformation on each position identically and separately.

\section{Double Path Networks}
\begin{figure*}[!t]
		\centering
		\includegraphics[width=0.95\columnwidth, height=65ex]{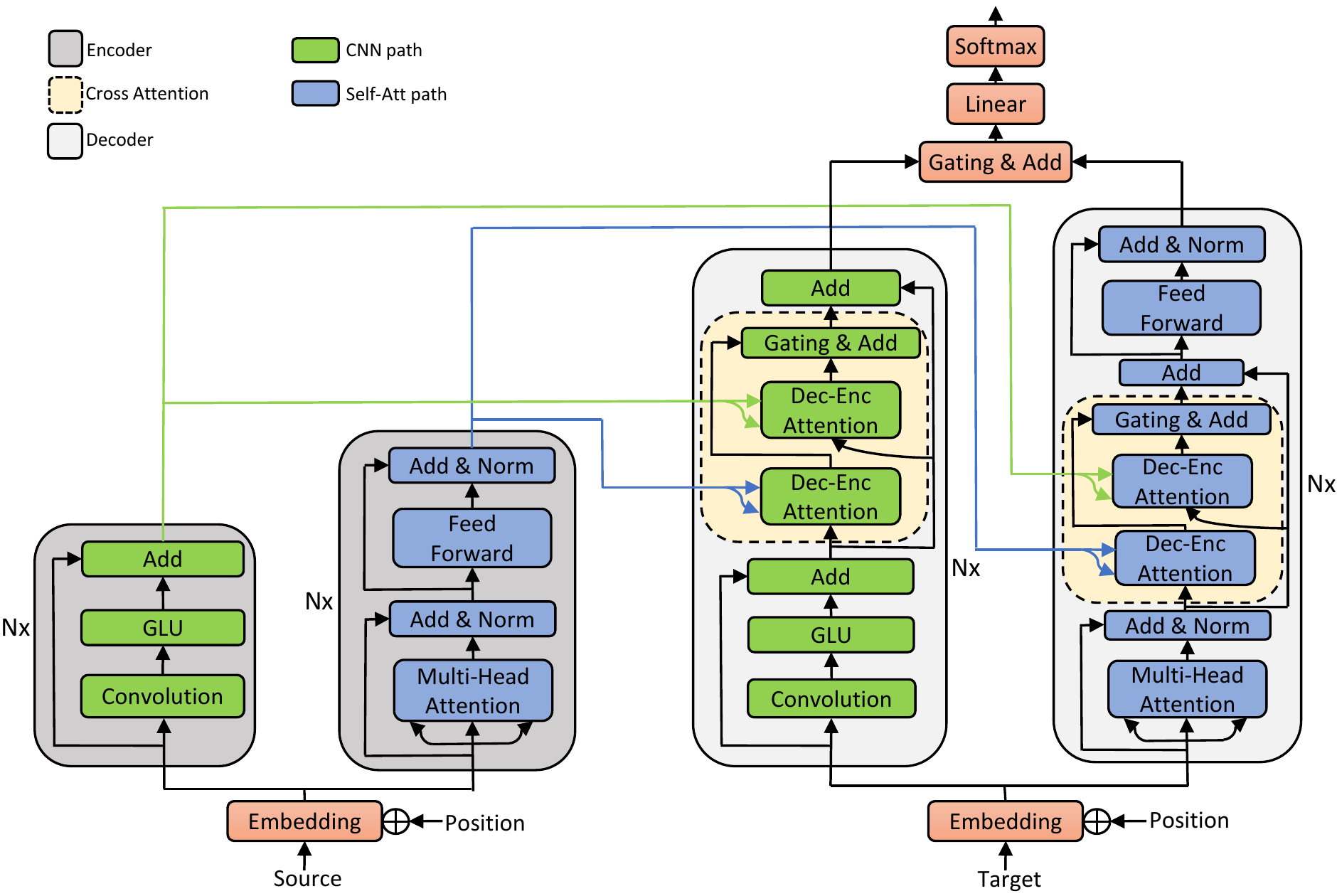}
		\caption{DPN-S2S architecture. The CNN and SAN path are depicted with green and blue module, respectively. The encoder part is colored with dark grey, attention fusion part with light yellow, and decoder part with light grey. The blue and green line show the information flow from encoder to decoder.}
		\label{fig:dcml}
	\end{figure*}

	In our work, we introduce double path networks which incorporate CNN and SAN for sequence to sequence learning. A detailed model structure is shown in Figure \ref{fig:dcml}. 
		
	\subsection{Double Path Encoder}
		To leverage the advantage of CNN and SAN layer, we design an encoder which includes a CNN and SAN path, introduced in Section 2.2 and 2.3 respectively. We sum up the position embedding $\textbf{p}=(p_1,...,p_m)$ and   word embedding $\textbf{w}=(w_1,...,w_m)$ as the input of encoder, to give the model a sense of the token's position in the sentence. For each path, we employ residual connections to stack multiple layers for deeper network. Consequently, we can collect two various context representations from encoder. 
	
		\subsection{Cross Attention with Gating}	 
		Between the encoder and the decoder, we develop a cross attention module with gating fusion, to automatically pick up the information needed to decode target word. As shown in Figure \ref{fig:dcml}, the attention fusion module consists of two submodules: $Dec-Enc\ Attention$ and $Gating\ \&\ Add$. We now give detailed descriptions of the two submodules.  
		
		Each path in the decoder take encoder's double-path outputs as context to make attention. Thus, there are four types of information flows through encoder to decoder. The $Dec-Enc\ Attention$ module in the decoder takes Eq. (\ref{eq:eq3}) to compute attention results, which is denoted as:
		\begin{equation}
		\begin{aligned}
		ctx^{cc} &= Attention(q^c, k^c, v^c), & ctx^{ca} = Attention(q^c, k^a, v^a),\\
		ctx^{ac} &= Attention(q^a, k^c, v^c), & ctx^{aa} = Attention(q^a, k^a, v^a).
		\end{aligned}
		\end{equation}
		For example, $ctx^{ca} \in \mathcal{R}^{d}$ means the attention result with attention query $q^{c}$ from CNN path in the decoder and attention key $k^{a}$ and value $v^{a}$ from SAN path in the encoder. 
		
		In order to fully exploit the information captured by different encoder paths, we design a gating mechanism to fuse them for balancing information usage between local connectivity and global semantics. This fusion gate is shown as the $Gating\ \&\ Add$ module in the right side of Figure \ref{fig:dcml}, formulated as follows:
		\begin{equation}
		\begin{aligned}
		g^c &= sigmoid([ctx^{cc}, ctx^{ca}] W^{c} + b^{c}),\\
		g^a &= sigmoid([ctx^{aa}, ctx^{ac}] W^{a} + b^{a}),\\
		ctx^{c} &= ctx^{cc} \cdot (1-g^c) + ctx^{ca} \cdot g^c,\\
		ctx^{a} &= ctx^{aa} \cdot (1-g^a) + ctx^{ac} \cdot g^a, 
		\end{aligned}
		\end{equation}
		where $g^{c}$ and $g^{a}$ are scalar attention gates for CNN context and SAN context; $[\cdot, \cdot]$ is an element concatenate operation; $W^{c}$ and $W^{a}\in \mathcal{R}^{2d}$ are the weight matrices; $b^{c}$ and $b^{a}$ are the scalar biases of the attention gates; and $ctx^{c}$ and $ctx^{a}$ are the fusion context vectors for the current CNN layer and SAN layer. For the gate $g^c$ and $g^a$, both the two attention results are used as inputs to the gating computation.
	
	\subsection{Double Path Decoder} 
		The decoder is illustrated in the right side of Figure \ref{fig:dcml}. It is similar to the encoder except for the decoder-to-encoder attention. Each layer in CNN and SAN contains two decoder-to-encoder attention modules, which is denoted in last subsection. We also need remove the future information in the convolution and self-attention operations in order to be consistent with decoding phase, for target sentence is generated sequentially, without any prior knowledge about next word.
	
		We also apply a gating mechanism to combine the outputs of the two decoder paths and feed the combined result into the $softmax$ layer to generate next word:
		\begin{equation}
			\begin{aligned}
				g^{o} &= sigmoid( [z^{c}, z^{a}] W^{o} + b^{o}),  \\
				z^{o} &= z^{c}\cdot(1-g^{o}) + z^{a} \cdot g^{o},\\
				P(y) &= softmax(z^{o} W^{s} + b^{s}),
			\end{aligned}
		\end{equation}
		where $z^{c}$ and $z^{a}$ are the outputs of CNN and SAN path, respectively. $z^{o}$ is the fusion output and $W^{s}$, $b^{s}$ are linear transform weights and biases before $softmax$.

\section{Experiment Settings}
	In this section, we introduce the datasets used in experiments, model settings, and training details.
	\subsection{Datasets}
	We evaluate our proposed method on two neural machine translation tasks and one abstractive text summarization task. The datasets are described as follows:		
	\paragraph{IWSLT2014 German-English} We use the dataset of IWSLT14 German-English machine translation track \cite{Cettolo2014Report} for training and evaluation. We tokenize the data which come from TED and TEDx talks. After preprocessing, the dataset remains 160K training sentences and 7K development sentences\footnote{https://github.com/facebookresearch/fairseq-py/blob/master/data/prepare-iwslt14.sh}. We concatenate dev2010, tst2010, tst2011 and tst2012 as the test set. We consider a joint source and target byte-pair encoding (BPE) with 10K types \cite{DBLP:conf/acl/SennrichHB16a}.
	
	\paragraph{Nist Chinese-English} We use the corpus consisting of 1.25M sentence pairs extracted from LDC corpora \footnote{The corpora includes LDC2002E18, LDC2003E07, LDC2003E14, Hansards portion of LDC2004T07, LDC2004T08 and LDC2005T06.}. This dataset contains 27.9M Chinese words and 34.5M English words respectively. We learn a 25K subwords dictionary based on BPE for source and target languages separately. After tokenization, we get a source vocabulary with 37K words and a target vocabulary  with 25K words. We choose Nist-2003 as the development set, and Nist-\{2004, 2005, 2006, 2008, 2012\} as the test sets.
	
	\paragraph{Text Summarization} We use the Gigaword corpus \cite{Graff03Eg} and follow the processing as Rush \shortcite{DBLP:conf/emnlp/RushCW15}. We obtain 3.8M training samples and 190K validation samples. We evaluate our approach on Gigaword dataset, which is a widely used test set with 2000 article-title pairs. Similar to Gehring \shortcite{DBLP:conf/icml/GehringAGYD17}, we use source and target vocabulary both with 30K words. Follow Gehring \shortcite{DBLP:conf/icml/GehringAGYD17}, we require the length of generated sequence is at least 14 words.
	
	\subsection{Model Settings}

	\paragraph{IWSLT14 German-English}
	For CNN path, we adopt 4 idential convolutional layers with 256 neurons in both the encoder and the decoder. As one self-attention layer is comprised of 2 sublayers, we set 2 layers for SAN to guarantee the depth consistency of network. The hidden size and the filter size of SAN block are set as 256 and 1024 respectively. The word embedding dimension is 256. And the dropout rate is set as 0.1.
	In order to demonstrate the efficiency of our method, we compare our method with four following baselines: (i) \emph{Deeper CNN Model}, a standard CNN based model with 8 layers of 256 neurons. (ii) \emph{Wider CNN Model}, a standard CNN based model with 4 layers of 512 neurons. (iii) \emph{Wider SAN Model}, a standard SAN based model with 2 layers of 512 neurons. (iv) \emph{Deeper SAN Model}, a standard SAN based model with 4 layers of 256 neurons. Besides, we also list some related works for comparison.
	
	\paragraph{Nist Chinese-English}
	We perform a deep model on Nist Chinese-English translation task. This model adopts 12 stacked layers for the CNN path and 6 stacked layers for SAN path in the encoder and decoder respectively. The hidden size of both CNN/SAN layers are both set as 256 and the filter size of SAN layer is 1024. We set 256 neurons for the word embedding. The dropout rate is set as 0.2. For the purpose of comparison, We choose two model settings as baselines: (i) \emph{CNN Model}, a CNN based model with 12 stacked layers of 512 neurons (ii) \emph{SAN Model}, a SAN model with 6 stacked layers of 512 neurons and the filter size of layer is 2048. These model settings contain approximately similar number of parameters in contrast to our model. We also list some previous works for comparison.
	
	\paragraph{Text Summarization}
	We follow the same model setting as described in IWSLT14 German-English translation task. For this task, We adopt the models including RNN MRT \cite{DBLP:journals/corr/AyanaSLS16}, WFE \cite{DBLP:journals/corr/SuzukiN17} and ConvS2S \cite{DBLP:conf/icml/GehringAGYD17} as the baselines. 
	
	\begin{wraptable}{r}{9cm}
		\centering
		\begin{tabular}{l c | r| r }
			\hline
			& & Params & BLEU \\	
			\hline \hline 
			AC+LL \cite{DBLP:journals/corr/BahdanauBXGLPCB16} & & - & 28.53 \\
			NPMT+LM \cite{DBLP:journals/corr/HuangWZD17} & & - & 29.16 \\
			\hline
			Wider CNN & & 20.37M & 30.63 \\
			Wider SAN & & 27.06M & 31.29 \\
			Deeper CNN & & 13.82M & 30.70 \\
			Deeper SAN & & 13.54M & 31.43 \\			
			\hline 
			DPN-S2S & &11.57M & \textbf{31.99} \\
			\hline
		\end{tabular}
		\caption{IWSLT14 German-English translation performance.}
		\label{de-en}
	\end{wraptable}
	
	\subsection{Training and Evaluation} We adopt Nesterov Accelerated Gradient (NAG) \cite{DBLP:journals/corr/abs-1212-0901} as training optimizer. The initial learning rate is set as 0.25 for IWSLT14 German-English and text summarization, and 0.50 for Nist Chinese-English, with a decay schedule that shrinking by 10 when the validation loss stops decreasing. Each training batch contains approximately 4000 source and target tokens. During inference, we set beam size as 5 for IWSLT14 German-English and text summarization, and 10 for Nist Chinese-English. All models are implemented in PyTorch based on \emph{fairseq-py}\footnote{https://github.com/facebookresearch/fairseq-py}. We use one Nvidia Titan x Pascal GPU for IWSLT14 German-English and text summarization, and 4 GPUs for Nist Chinese-English.

\section{Results}
	In this section, we report our results on translation and summarization tasks. In addition, we also conduct  a series of extensive analyses for a better understanding of our model.
		
	\subsection{IWSLT2014 German-English} We evaluate the translation accuracy by BLEU \footnote{\url{https://github.com/moses-smt/mosesdecoder/blob/master/scripts/generic/multi-bleu.perl}} \cite{DBLP:conf/acl/PapineniRWZ02}. Table \ref{de-en} shows the experiment results on German-English translation task. We list the results of the wider network, deeper network and some related works for a comparison. The results can be summarized as follow: 
	
	When compared with wider network, DPN-S2S achieves an improvement of 1.36 BLEU and 0.70 BLEU over CNN and SAN model. When compared with deeper network, DPN-S2S achieves an improvement of 1.29 BLEU and 0.56 BLEU over CNN and SAN model. Generally, we note that DPN-S2S can produce better performance than wider or deeper networks with fewer parameters. In addition, we choose two methods which used an actor-critic algorithm \cite{DBLP:journals/corr/BahdanauBXGLPCB16} and 
	a phrased-based model with an auxiliary language model \cite{DBLP:journals/corr/HuangWZD17} separately. We find our method also outperforms these approaches. All of these comparsions indicate the effectiveness of our method.
	
	\begin{table}
		\centering
		\begin{tabular}{l | c c c c c }
			\hline
			& Nist04 & Nist05 & Nist06 & Nist08 & Nist12 \\
			\hline \hline 
			MC-NMT \cite{zh-en_1} &40.79 & 38.49 & - & 31.51 & 26.90\\
			NMT + Distortion \cite{zhang-EtAl:2017:Long3}  &40.52 & 36.81 & 35.77 & - & - \\
			SD-NMT \cite{DBLP:journals/corr/abs-1711-04231} &39.81 & 36.74 & 34.63 & 28.61 & - \\
			\hline
			SAN & 40.39 & 40.38 &38.06& 31.67 & 30.32 \\
			CNN & 40.80 & 38.16 &37.89& 30.70 & 29.55 \\
			\textbf{DPN-S2S} & \textbf{41.26} & \textbf{41.12} &\textbf{38.87}& \textbf{32.66} & \textbf{31.17} \\
			\hline
		\end{tabular}
		\caption{Performance of different systems on Nist Chinese-English translation task.}
		\label{zh-en}
	\end{table}
	
	\subsection{Nist Chinese-English}
	Table \ref{zh-en} lists the results about our model on each Nist Chinese-English test set, together with several NMT baselines. These baselines are RNNSearchs with some well acknowledged techniques including: 1) Using multi-channel information into encoder (MC-NMT) \cite{zh-en_1}; 2) Incorporating word reorder information into NMT (NMT+Distortion) \cite{zhang-EtAl:2017:Long3}; 3) An attention mechanism with a directed-syntax (SD-NMT) \cite{DBLP:journals/corr/abs-1711-04231}. 
	We can observe that our method surpasses SAN and CNN baseline by 0.74-0.99 and 0.46-2.96 BLEU score in different test sets. 
	In addition, we also compare our results with some related works, and our model outperforms MC-NMT by 0.47-4.27 BLEU, NMT+Distortion by 0.74- 4.31 BLEU and SD-NMT by 1.44-4.38 BLEU in Nist test sets. 
	These results on Nist Chinese-English dataset prove that our model can also achieve improvement in large dataset.
	
	\begin{table}[h]
		\centering
		\begin{tabular}{l | c  c  c }
			\hline
			& \multicolumn{3}{c}{Gigaword} \\
			& RG-1 (F) & RG-2 (F) & RG-L (F) \\
			\hline
			RNN MRT \cite{DBLP:journals/corr/AyanaSLS16}  & 36.54 & 16.59 & 33.44 \\ %
			WFE \cite{DBLP:journals/corr/SuzukiN17} & 36.30 & 17.31 & 33.88 \\
			ConvS2S \cite{DBLP:conf/icml/GehringAGYD17} & 35.88 & 17.48 & 33.29 \\ %
			\hline
			DPN-S2S & \textbf{36.92} & \textbf{17.91} & \textbf{34.32} \\
			\hline
		\end{tabular}
		\caption{Accuracy on Gigaword text summarization task in terms of ROUGE-1 (RG-1), ROUGE-2 (RG-2), and ROUGE-L (RG-L). F stands for F1-score.}
		\label{text}
	\end{table}
	
	\subsection{Text Summarization} 	We employ three variants of ROUGE \cite{Lin:2004} as evaluation metrics for text summarization: ROUGE-1 (unigrams), ROUGE-2 (bigrams), and ROUGE-L (longest-common substring). 
	\begin{wraptable}{r}{8cm} 
		\centering
		\begin{tabular}{c| c c | c c | c}
			\hline
			\multirow{3}{*}{ID} & \multicolumn{2}{c|}{Encoder} & \multicolumn{2}{c|}{Decoder} & \multirow{3}{*}{BLEU}\\		
			& \multirow{1}{*}{CNN} & \multirow{1}{*}{SAN} &  \multirow{1}{*}{CNN} & \multirow{1}{*}{SAN} &  \\
			\hline \hline
			M1 & $\Delta$ & & $\Delta$ & & 30.38 \\ 
			\hline
			M2 & $\Delta$ & & & $\Delta$ & 31.00 \\ 
			\hline
			M3 & $\Delta$ & & $\Delta$ & $\Delta$ & 31.26 \\
			\hline
			M4 & & $\Delta$ & $\Delta$ & & 29.80 \\
			\hline
			M5 & & $\Delta$ & & $\Delta$ & 30.63 \\ 
			\hline
			M6 & & $\Delta$ & $\Delta$ & $\Delta$ & 31.11 \\ 
			\hline
			M7 & $\Delta$ & $\Delta$ & $\Delta$ & & 31.03 \\
			\hline
			M8 & $\Delta$ & $\Delta$ & & $\Delta$ & 31.43 \\ 
			\hline
			M9 & $\Delta$ & $\Delta$ & $\Delta$ & $\Delta$ & \textbf{31.99} \\
			\hline
		\end{tabular}
		\caption{Ablation studies of DPN-S2S on IWSLT14 German-English. $\Delta$ means the model contains this component.}
		\label{tab:application}
	\end{wraptable}
	We compare our model with some previous work including RNN MRT \cite{DBLP:journals/corr/AyanaSLS16} which adopts RNNSearch on this task,  WFE \cite{DBLP:journals/corr/SuzukiN17} which addresses the problem of repeated sequence generation, and ConvS2S \cite{DBLP:conf/icml/GehringAGYD17}. Table \ref{text} shows the text summarization results on the Gigaword dataset. 
	DPN-S2S outperforms all the three models by 0.43-1.04 ROUGE-1, 0.43-1.32 ROUGE-2 and 0.64-1.23 ROUGE-L in terms of F1 score.
	
	\subsection{Ablation Study}
	To understand to what extent each path in the encoder/decoder can affect the model performance, we conduct some experiments for ablation study. For conciseness we just choose one task (IWSLT14 German-English) for this study. Table \ref{tab:application} shows the experiment settings and results. All the models are of the same dimensions for hidden states and word embedding. 
	Note that model M9 is our proposed DPN-S2S. We have the following observations.
	\begin{itemize}
		\item Comparing model M4 with M7, M5 with M8, and M6 with M9, we can see that adding the CNN encoder leads to 0.80-1.23 BLEU gain. 
		\item Comparing model M2 with M3, M5 with M6, and M8 with M9, we observe that adding the CNN decoder leads to 0.26-0.56 BLEU gain. 
		\item Comparing model M1 with M7, M2 with M8, and M3 with M9, we find that adding the SAN  encoder results in 0.43-0.73 BLEU improvement. 
		\item Similarly comparing model M1 with M3, M4 with M6, and M7 with M9, we see 0.88-1.31 BLEU gain by adding the SAN decoder. 
	\end{itemize}
	These results suggest that both the CNN path and the SAN path make positive contributions to the overall architecture, and they are needed in both the encoder and decoder.
	
	\subsection{Attention Visualization}
	
	\begin{wraptable}{r}{8.5cm}
		\centering
		\begin{tabular}{c | c | c }
			\hline
			\diagbox{Encoder}{Decoder} & CNN & SAN \\
			\hline
			CNN & 2.208 & 1.406 \\
			\hline
			SAN & 1.886 & 1.377 \\
			\hline
		\end{tabular}
		\caption{The entropy of the alignments of CNN/SAN decoder to CNN/SAN encoder on IWSLT14 German-English.}
		\label{distribution}
	\end{wraptable}
	
	To demonstrate the different characteristics of the CNN and SAN path, we analyze the alignments (attention coefficients) from the four types of decoder-to-encoder attention. The alignments are computed by $Softmax(qk^T)$ from Eq. (\ref{eq:eq3}), and the alignments from each token in decoder form a distribution. We use the entropy of the distribution $-\sum_{i}x_{i}log x_{i}$ to depict to what extent the target token focuses on the source tokens. Small entropy means less uncertainty, i.e., the attention is more concentrated. 

	We choose the 6750 sentences in the German-English test set and calculate the entropy of each target token's alignments distribution, averaged by target token in the sentence and then averaged by sentence, as shown in Table \ref{distribution}. The entropy of alignments from both paths in decoder to CNN path in encoder (the first row of the table) is larger than that of SAN path in encoder. This is consistent with our intuitions: CNN focuses more on local information, and so the decoder needs to attend to multiple hidden states of the CNN path in the decoder to extract necessary information for target word generation; in contrast, each hidden state in the SAN path has already included global semantics, and so the decoder only needs to focus on a small number of hidden states of the SAN path in the encoder, resulting in more focused alignments and smaller entropy (the second row of the table).

\begin{figure*}[!t] 
		\centering
		\begin{subfigure}[h]{0.22\textwidth}
			\centering
			\includegraphics[width=\textwidth]{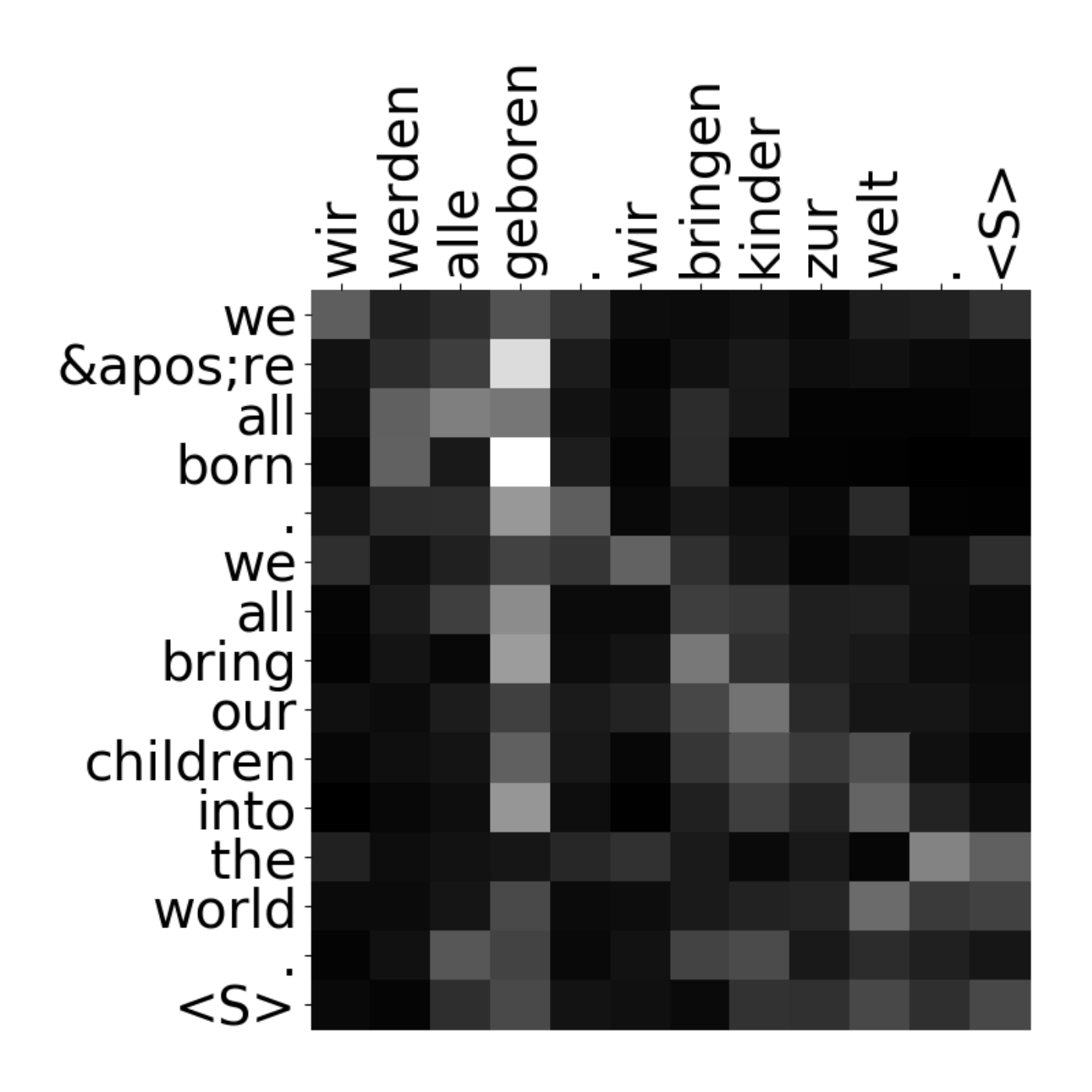}
			\caption{CNN to CNN}
			\label{sub1}
		\end{subfigure}
		\begin{subfigure}[h]{0.22\textwidth}
			\centering
			\includegraphics[width=\textwidth]{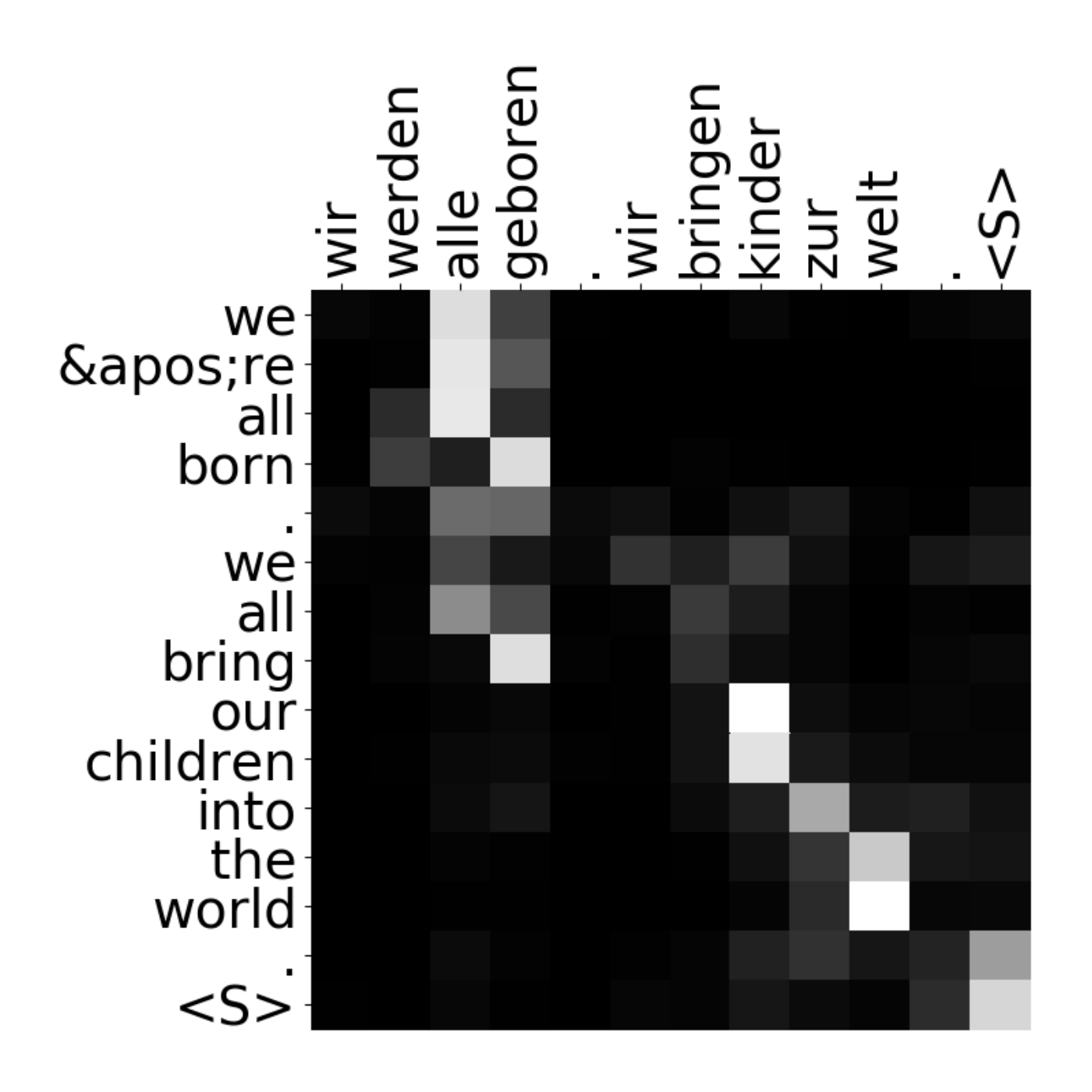}
			\caption{CNN to SAN}
			\label{sub2}
		\end{subfigure}
		\begin{subfigure}[h]{0.22\textwidth}
			\centering
			\includegraphics[width=\textwidth]{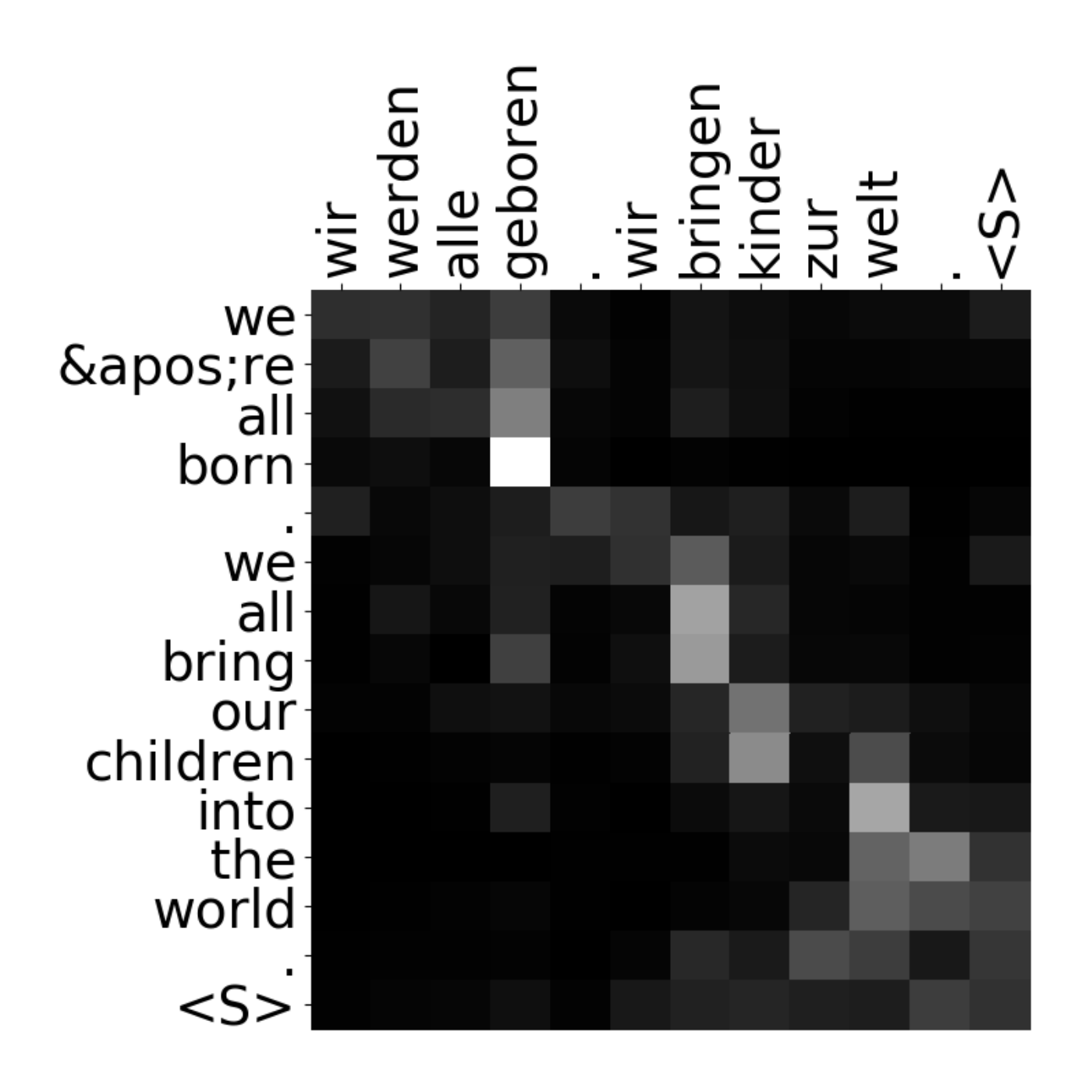}
			\caption{SAN to CNN}
			\label{sub3}
		\end{subfigure}
		\begin{subfigure}[h]{0.22\textwidth}
			\centering
			\includegraphics[width=\textwidth]{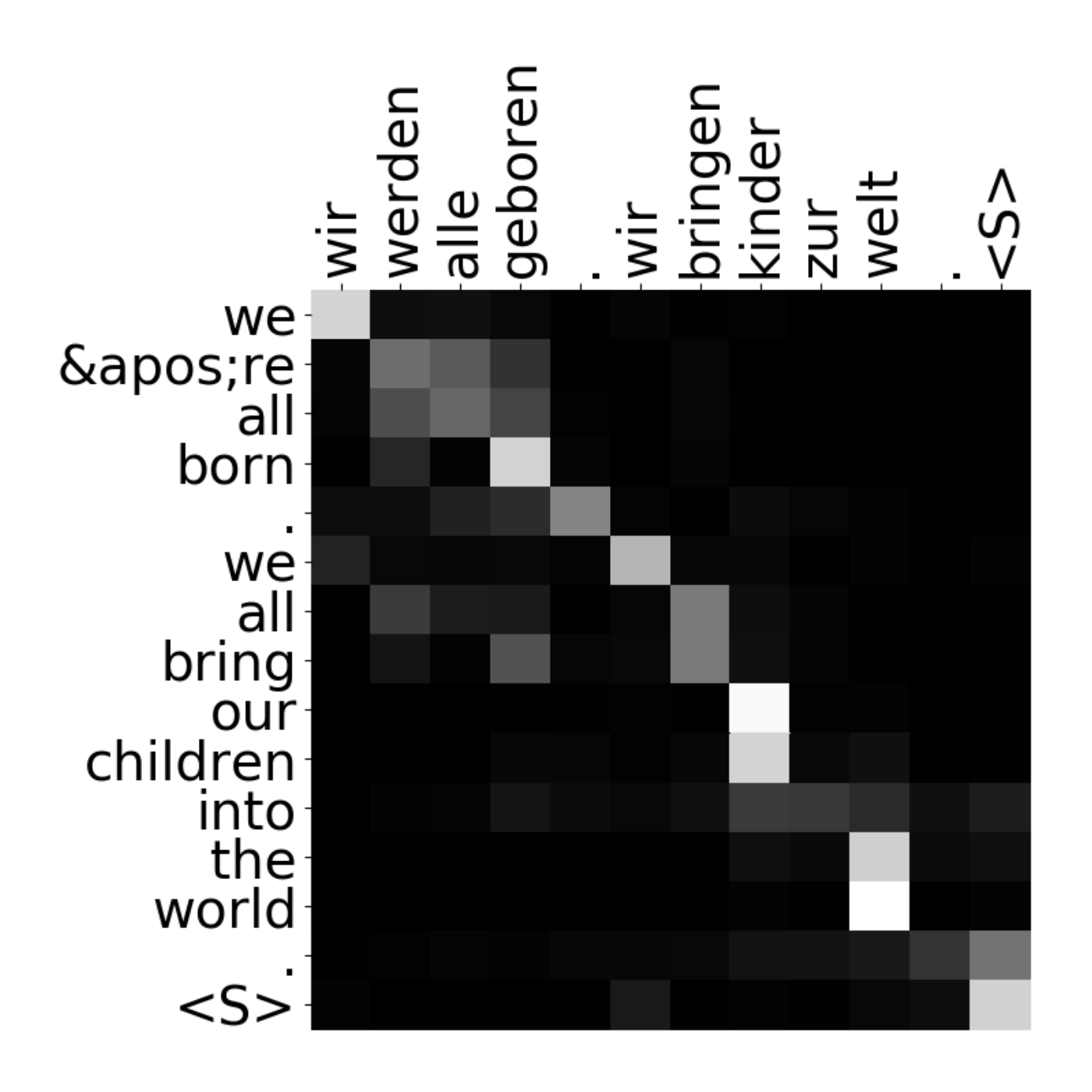}
			\caption{SAN to SAN}
			\label{sub4}
		\end{subfigure}
		\caption{The alignments of CNN/SAN decoder to CNN/SAN encoder from DPN-S2S. The x-axis and y-axis of each plot correspond to the words in the source language and target language. Each pixel shows the alignment scores of the target word to source word.}
		\label{alignment_score}
	\end{figure*}

	For an intuitive understanding, we visualize the alignments of a sentence pair in Figure \ref{alignment_score}. Comparing Figure \ref{sub2} with \ref{sub1}, we can see that the attention from the CNN path of the decoder to the SAN path of the encoder is more focused than that to the CNN path of the encoder. The same phenomenon can be found for attention from the SAN path of the decoder to the encoder, by comparing Figure \ref{sub4} with \ref{sub3}.

	\begin{table}
		\centering
		\begin{tabular}{l r l}
			\toprule
			& & Translation \\
			\cmidrule{1-1} \cmidrule{3-3}
			Source (De) &  &\fontsize{9pt}{10} \selectfont das ist die linke hälfte, die die logische seite ist und dann die rechte h\"{a}lfte, die die intuitive seite ist. \\
			\cmidrule{1-1} \cmidrule{3-3}
			Target (En) & & \fontsize{9pt}{10} \selectfont there 's \emph{the left half}, which is the logical side,  and then \emph{the right half}, which is the intuitive. \\
			\cmidrule{1-1} \cmidrule{3-3}
			CNN & & \fontsize{9pt}{10} \selectfont this is \emph{half the left half}, which is the logical side, and then \emph{half the rights} that is the intuitive side. \\
			\cmidrule{1-1} \cmidrule{3-3}
			SAN & & \fontsize{9pt}{10} \selectfont that's the left half, which is the logical side , and then \emph{half}, that 's the intuitive side. \\
			\cmidrule{1-1} \cmidrule{3-3}
			DPN-S2S & & \fontsize{9pt}{10} \selectfont that's \emph{the left half}, which is the logical side,  and then \emph{the} \emph{right half}, which is the intuitive side. \\
			\bottomrule
		\end{tabular}
		\caption{A German-English translation case to demonstrate DPN-S2S outperforms CNN and SAN in terms of translation accuracy.}
		\label{de-en_cases}
	\end{table}
	
	\subsection{Case Study}
	To better understand the advantage of DPN-S2S over single path models, Table \ref{de-en_cases} shows a case from the German-English test set. We list the generated target sentence by CNN based model, SAN based model and DPN-S2S, respectively. As can be seen, CNN based model generates detailed words such as ``half the left half'' and ``half the rights'', without a global view to correctly organize these local meanings. SAN based model misses the local details about the ``the right half'' information. Our DPN-S2S well combine the local information from the CNN based path and the global information from the SAN based path, resulting in a better translation. 

\section{Conclusion}
	In this paper, we have proposed double path networks for sequence to sequence learning, named \emph{DPN-S2S}, which uses a cross attention module to leverage the advantages of two different models, and achieves the state-of-art performance in several sequence to sequence tasks.

	We plan to explore the following directions in the future. First, we will test the new model on more language pairs for neural machine translation and other sequence to sequence tasks. Second, it is interesting to investigate how to design better structures for information fusion. Third, we would like to extend the current double path model to multiple paths.
	
\newpage
\section*{Acknowledgments}
This work was supported by The National Key Research and Development Program of China under Grant 2018YFB1004904.

\bibliographystyle{acl}
\bibliography{coling2018}

\end{document}